\documentclass{article}
\usepackage{epsfig}
\usepackage{amssymb}
\usepackage{latexsym}

\usepackage{color}

\begin{document}

\title{
Bilateral Spatial Reasoning about Street Networks:
Graph-based RAG with Qualitative Spatial Representations}
\author{R. Moratz, N. Daute, J. Ondieki, 
M. Kattenbeck, M. Krajina, I. Giannopoulos}

\maketitle

\begin{abstract}
This paper deals with improving the capabilities of Large Language Models (LLM) to provide route instructions for pedestrian wayfinders by means of qualitative spatial relations. We use a method called Retrieval-Augmented Generation (RAG). RAG supports the LLM with context information based on the specific query. We assess the impact the added information has on model performance for generating pedestrian route instructions. Our findings encourage further integration of qualitative spatial data into LLM applications—potentially benefiting areas such as digital navigation aids, smart city tools, and accessibility technologies.
\end{abstract}

\section{Introduction}
\begin{figure}[h]
\begin{center}
\includegraphics[width=10cm]{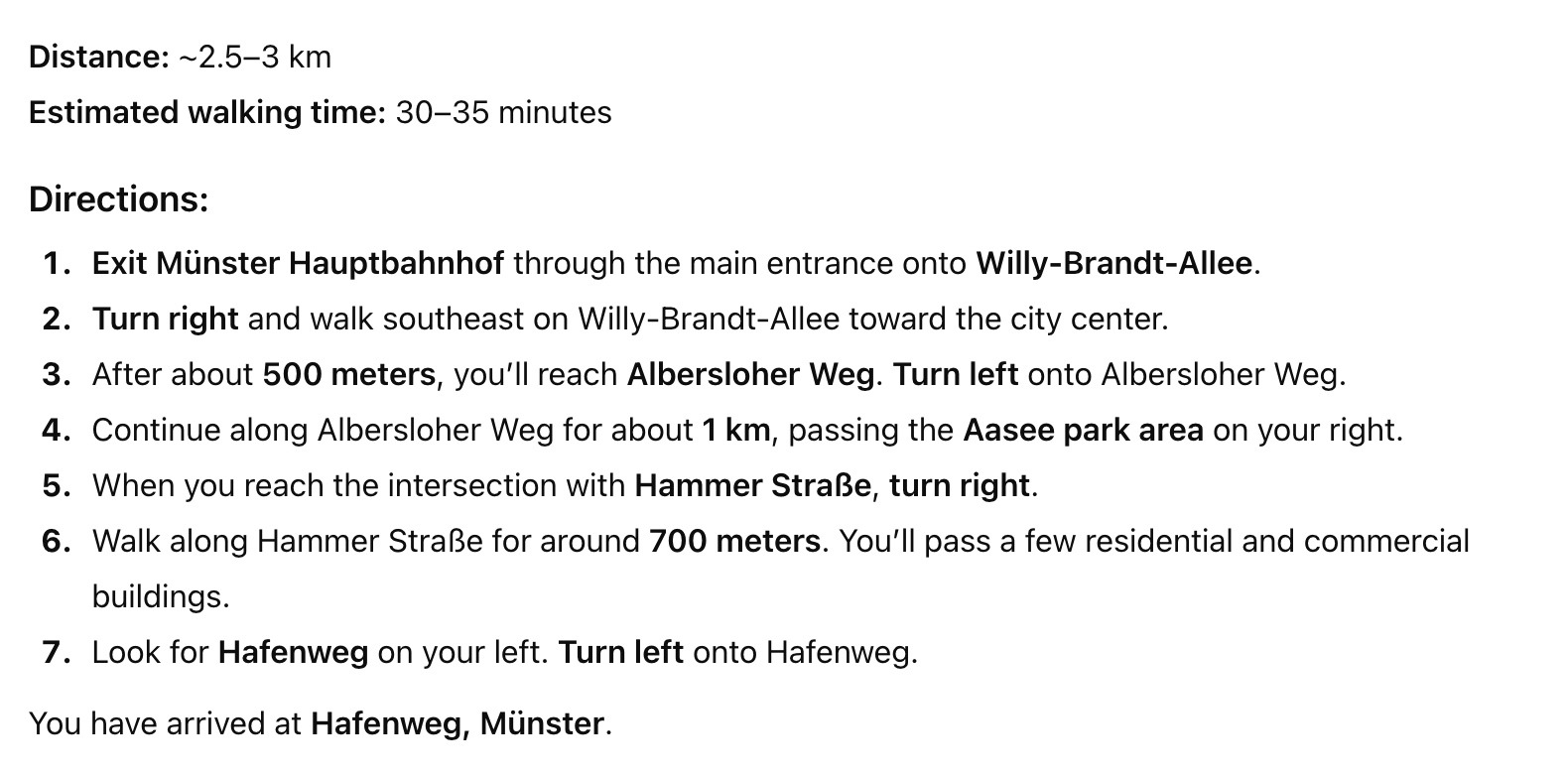}
\caption{\label{sampleNav} Route instruction generated by LLM}
\end{center}
\end{figure}

Up until now, Large Language Models show rather weak performance in providing route instructions to pedestrian wayfinders. 
For example we can ask an LLM (ChatGPT-4o by OpenAI) for the direction from M{\"u}nster central station 
to nearby street Hafenweg (see figure\ref{sampleNav}).

\begin{figure}[h]
\begin{center}
\includegraphics[width=10cm]{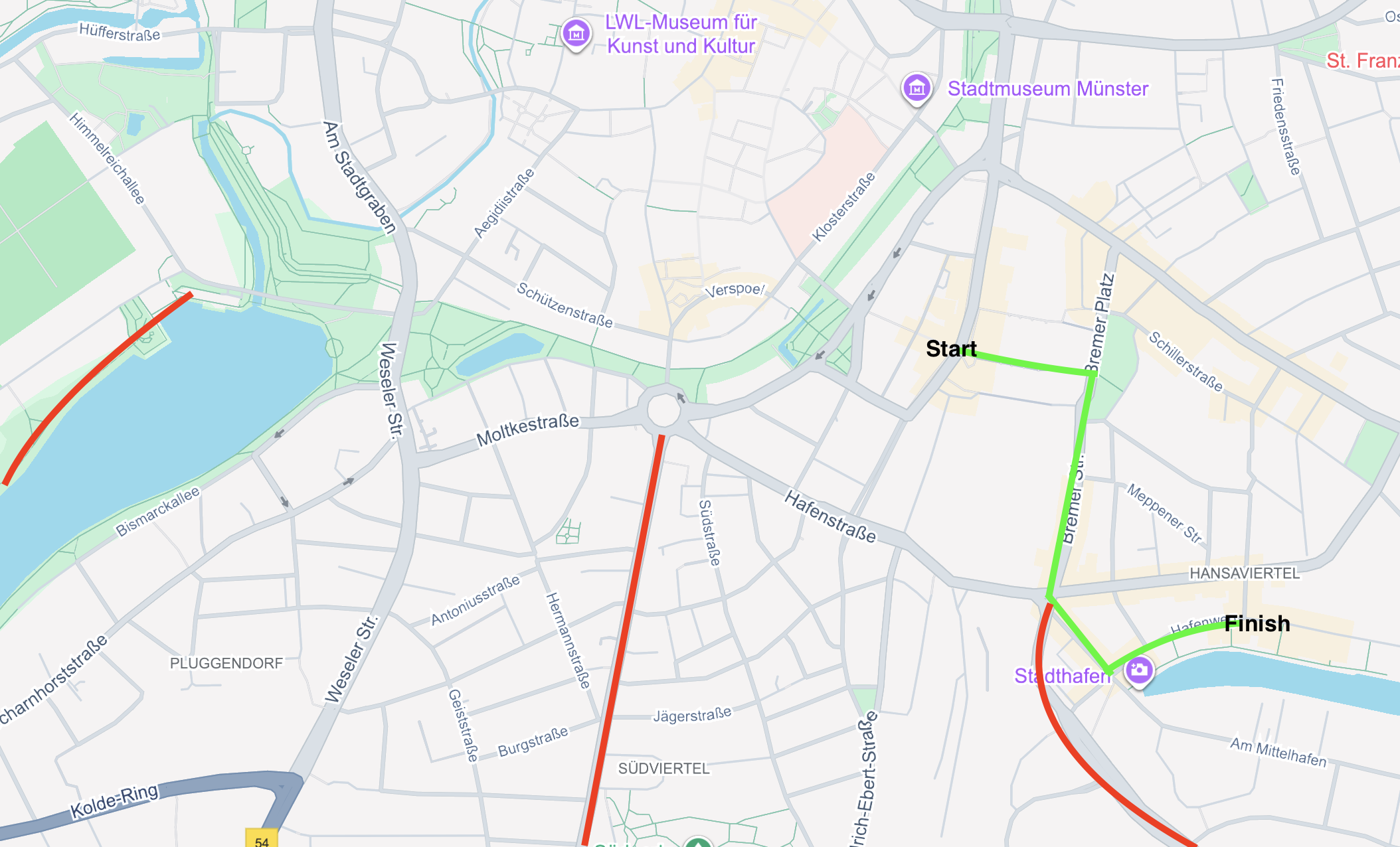}
\caption{\label{mapNav} Map detail showing erroneous route instruction}
\end{center}
\end{figure}

The model responded quickly with a seemingly convincing set of directions for the given problem.
However, upon closer inspection, it becomes evident that LLM navigation is heavily prone to hallucinations and other errors.
In the first step, the model suggests leaving the central station onto "Willy-Brandt-Allee", which is a street that does not exist in Münster.
Subsequently, the model suggests walking toward the city center, while the Hafenweg is actually located away from the city center.
Next, turning onto the "Albersloher Weg" is suggested.
While this road exists in Münster and is close to the actual destination, it does not make sense to take the road for the given navigation task.
The same can be said for the following two steps, where turning onto "Aaseepark" and "Hammer Straße" is suggested, with neither of them being sensible choices for the problem, while also being completely disconnected, making navigation along the suggested route impossible.
Finally, the model suggests turning onto the destination street "Hafenweg", which is the correct destination, but impossible given the previous incorrect steps.
When comparing this route proposed by ChatGPT to a route provided by Google Maps, the problem becomes even more evident.
In the figure, the correct route from "Start" to "Finish" is shown in green:
A few simple turns from the Central Stations is all it takes to reach the destination.
The generated LLM route displayed in red on the map is however far from correct:
Mutltiple disconnected segments show a complete lack of understanding the actual navigation task, with only one of them being close to the actual route.
Observations like these highlight the current limitations of LLMs in navigation performance, and motivated us to find approaches to boost their capabilities in this domain.

\clearpage

This paper deals with improving the capabilities of Large Language Models to provide route instructions for pedestrian wayfinders by means of qualitative spatial relations. We use a method called Retrieval-Augmented Generation (RAG). RAG supports the LLM with context information based on the specific query. 
For this purpose we use a qualitative spatial representation framework which is based on
oriented line segments (dipoles) as basic entities \cite{MoratzEtAlAIJ}. In our context a {\it qualitative} representation provides mechanisms
which characterize central essential properties of objects or
configurations.
A {\it quantitative} representation in contrast establishes a
measure in relation to
a unit of measurement which has to be generally available.
Qualitative spatial spatial representations usually deal with elementary objects (e.g.,
positions, directions, regions) and qualitative relations between them
(e.g., ''adjacent'', ''on the left of'', ''included in'').

\section{Graph-Based Retrieval Augmented Generation (Graph-RAG)}

Graph-based Retrieval Augmented Generation represents an advanced evolution of traditional RAG systems that leverages knowledge graphs to enhance information retrieval and generation quality \cite{GraphRAG}. With Graph-RAG language model receives not just text chunks, but structured information including entity mentions and their properties, explicit relationships between entities, graph paths showing logical connections, and multi-level abstractions (detailed facts and high-level summaries).

Traditional RAG retrieves relevant text chunks from a vector database using semantic similarity, then feeds these chunks to a language model for generation. Graph-RAG enhances this by organizing information in a knowledge graph structure, where entities are nodes and relationships are edges, enabling more sophisticated retrieval strategies.

The system first constructs a knowledge graph from source documents by extracting entities (people, places, concepts, events), identifying relationships between entities, creating nodes for entities and edges for relationships, and storing both structural information and associated text content. Retrieved information is assembled into context by combining node content (entity descriptions), including edge information (relationships), incorporating graph structure (how entities connect), and adding hierarchical summaries when relevant. 

\section{Representation of Dipole Relations \label{basic}}

\begin{figure}[h]
\begin{center}
\includegraphics[width=5cm]{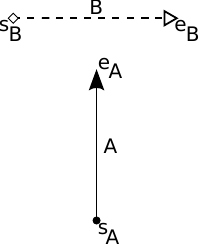}
\caption{\label{left} A dipole is an oriented straight line segment with start and end points}
\end{center}
\end{figure}

The basic entities we are using are dipoles, i.e.\ oriented line segments
formed by a pair of two points, a start point and an end
point. Dipoles are denoted by $A,
B, C,\ldots$, the start point by ${\bf s}_A$, the end point by ${\bf e}_A$,
respectively (see Figure~\ref{left}).

These dipoles are used for representing spatial objects with an
intrinsic orientation. Given a set of dipoles it is possible to
specify many different relations of different arity, e.g., depending
on the length of dipoles, on the angle between different dipoles, or on the
dimension and nature of the underlying space.
The goal of identifying different relations is to obtain a set of
jointly exhaustive and pairwise disjoint {\em atomic} relations such that
between any two dipoles exactly one relation holds. If these relations
form a algebra which fulfills certain requirements
it is possible to apply standard constraint-based reasoning
mechanisms which were originally developed for temporal reasoning 
and
which have also proved valuable for spatial reasoning.
In order to enable efficient reasoning, it should be tried to keep the
number of different base relations relatively small.

For this reason, we will restrict for now to using two-dimensional
continuous space, in particular ${\mathbb{R}}^2$, and distinguish the location and
orientation of the different dipoles only according to whether a point
lies to the left, to the right, or on the straight line through the
referring dipole.  Then ${\bf s}_B$ can either lie to the
left of $A$ (see figure \ref{left}),
on the straight line through $A$
or to the right of $A$, expressed as $A \;{\rm l}\; {\bf s}_B$,
$A \;{\rm o}\; {\bf s}_B$ or $A \;{\rm r}\; {\bf s}_B$, respectively.
Using these three relations between a dipole and a point it is
possible to specify the relations between two dipoles with the
following four relationships:
\begin{displaymath}
  A \;{\rm R}\; {\bf s}_B \wedge A \;{\rm R}\; {\bf
e}_B \wedge B \;{\rm R}\; {\bf s}_A \wedge B \;{\rm R}\; {\bf e}_A,
\end{displaymath}
where {\rm R} is one of $\{\rm r,l,o\}$.
Since this still leads to a large number of different atomic
relations, we require in the first version of our qualitative straight
line segment relation representation 
all points to be in {\em general position},
i.e., no more than two points are on a straight line. 
This gives us the following 14 relations (Please note that in this very first representation
all the four
points ${\bf s}_B,  {\bf e}_B,  {\bf s}_A,  {\bf e}_A$ have different locations.):
\begin{eqnarray*}
A \;{\rm rrrr}\; B & := & A \;{\rm r}\; {\bf s}_B \wedge A \;{\rm r}\; {\bf
e}_B \wedge B \;{\rm r}\; {\bf s}_A \wedge B \;{\rm r}\; {\bf e}_A \\
A \;{\rm rrrl}\; B & := &  A \;{\rm r}\; {\bf s}_B \wedge A \;{\rm r}\;
{\bf e}_B \wedge B \;{\rm r}\; {\bf s}_A \wedge B \;{\rm l}\; {\bf e}_A \\
& \vdots & \\
A \;{\rm llll}\; B & := &  A \;{\rm l}\; {\bf s}_B \wedge A \;{\rm l}\;
{\bf e}_B \wedge B \;{\rm l}\; {\bf s}_A \wedge B \;{\rm l}\; {\bf e}_A \\
\end{eqnarray*}

The cases
$A \;{\rm r}\; {\bf s}_B \wedge A \;{\rm l}\; {\bf e}_B \wedge B \;{\rm
r}\; {\bf s}_A \wedge B \;{\rm l}\; {\bf e}_A$ and
$A \;{\rm l}\; {\bf s}_B \wedge A \;{\rm r}\; {\bf e}_B \wedge B \;{\rm
l}\; {\bf s}_A \wedge B \;{\rm r}\; {\bf e}_A$
cannot be realized on the plane. 
These 14 relations between line segments were already
derived by Schlieder \cite{schlieder95}.

\begin{figure}[tb]
\begin{center}
\includegraphics[width=12cm]{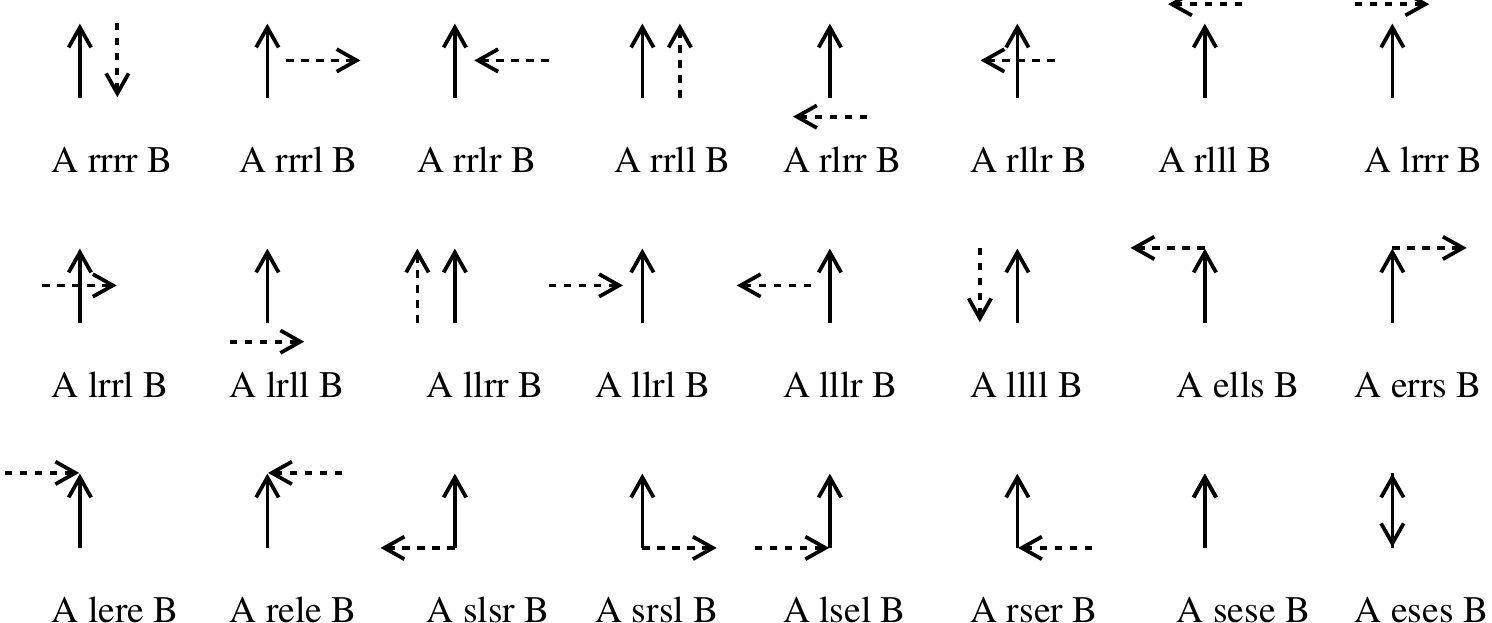}
\caption{\label{atomic_rel} The 24 atomic relations of the coarsest dipole calculus}
\end{center}
\end{figure}

In order to obtain an algebra, we need an identity relation.
This means in addition to \emph{left} or \emph{right}, we need 
configurations where more than one point have the same position on the two dimensional 
continuous space, $\mathbb{R}^{2}$. 
Therefore the smallest sensible dipole algebra has
two additional relative positions of start and end points,
namely the \emph{start}, or \emph{end} point of the other dipole.
Then ${\bf s}_B$ can be equivalent to the start point of $A$ or to the
end point of $A$. This is denoted as $A \;{\rm s}\; {\bf s}_B$ or
$A \;{\rm e}\; {\bf s}_B$, respectively.
Using these additional dipole-point relations, we obtain the following
ten additional dipole-dipole relations: $\{
{\rm ells},
{\rm errs},
{\rm lere},
{\rm rele},
{\rm slsr},
{\rm srsl},
{\rm lsel},
{\rm rser},
{\rm sese},
{\rm eses}
\}$.
Altogether we obtain 24 different atomic relations (see figure \ref{atomic_rel}). These relations
are jointly exhaustive and pairwise disjoint provided that all point positions are
in general position. The relation {\rm sese} is the requested identity
relation. 
We use ${\cal D}_{c}$ ('c' stands for 'coarse') to refer to the set of 24 atomic relations, and 
${\cal DRA}_{c}$ to refer to the powerset of ${\cal D}_{c}$ which
contains all $2^{24}$ possible unions of the atomic relations.

The relations which are introduced above in an informal way can be
defined in an algebraic way.
Every dipole $D$ on the plane
${\mathbb{R}}^2$ is an ordered pair
of two points ${\bf s}_D$ and ${\bf e}_D$, each of them is represented by its
Cartesian coordinates
$x$ and $y$, with
$x, y \in {\mathbb{R}}$ and
${\bf s}_D \not= {\bf e}_D$.
\begin{displaymath}
D = \left( {\bf s}_D, {\bf e}_D \right) ,   \qquad
{\bf s}_D  = \left( ({\bf s}_D)_x , ({\bf s}_D)_y \right)
\end{displaymath}

The basic relations are
then described
as equations with the coordinates as variables. The set of
solutions for a
system of equations describes all the possible coordinates for these points. 
A first such specification was presented in Moratz et. al. \cite{Moratz00_QSRwithLineSegs}. 

\subsection{Extended Versions of the Dipole Calculus}

\begin{figure}[thb]
\begin{center}
\includegraphics[height=2.1cm]{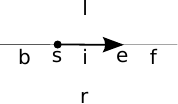}
\caption{\label{fine_relations} Extended dipole-point relations}
\end{center}
\end{figure}

In many domains we might want to represent spatial arrangements
in which more than two start or end points of dipoles are on a
straight line. 
Then we need three more dipole-point relations. The additional relations
describe the cases when the point is straight behind the dipole ($\rm b$),
in the interior of the dipole ($\rm i$) or straight in front of the dipole
($\rm f$). 
The corresponding regions are shown on Figure~\ref{fine_relations}.
Such a set of relations was proposed by Freksa \cite{freksa92b}.

Using the same notation scheme like the one for the coarse Dipole Relation
Algebra ${\cal DRA}_{c}$ we get the following 72 atomic relations:
\{
rrrr,
rrrl,
rrlr,
rrll,
rlrr,
rllr,
rlll,
lrrr,
lrrl,
lrll,
llrr,
llrl,
lllr,
llll,
ells,
errs,
lere,
rele,
slsr,
srsl,
lsel,
rser,
sese,
eses,
lllb,
llfl,
llbr,
llrf,
lirl,
lfrr,
lril,
lrri,
blrr,
irrl,
frrr,
rbrr,
lbll,
flll,
brll,
rfll,
rlli,
rrlf,
illr,
rilr,
rrbl,
rlir,
rrfr,
rrrb,
ffbb,
efbs,
ifbi,
bfii,
sfsi,
beie,
ffbb,
bsef,
biif,
iibf,
sisf,
iebe,
ffff,
fefe,
fifi,
fbii,
fsei,
ebis,
iifb,
eifs,
iseb,
bbbb, 
sbsb, 
ibib 
\}.
The derived fine grain Dipole Relation Algebra is called ${\cal DRA}_{f}$.

\clearpage

\subsection{Representation of street networks in the dipole calculus \label{streetnetworks}}

\begin{figure}[h]
\begin{center}
\includegraphics[width=10cm]{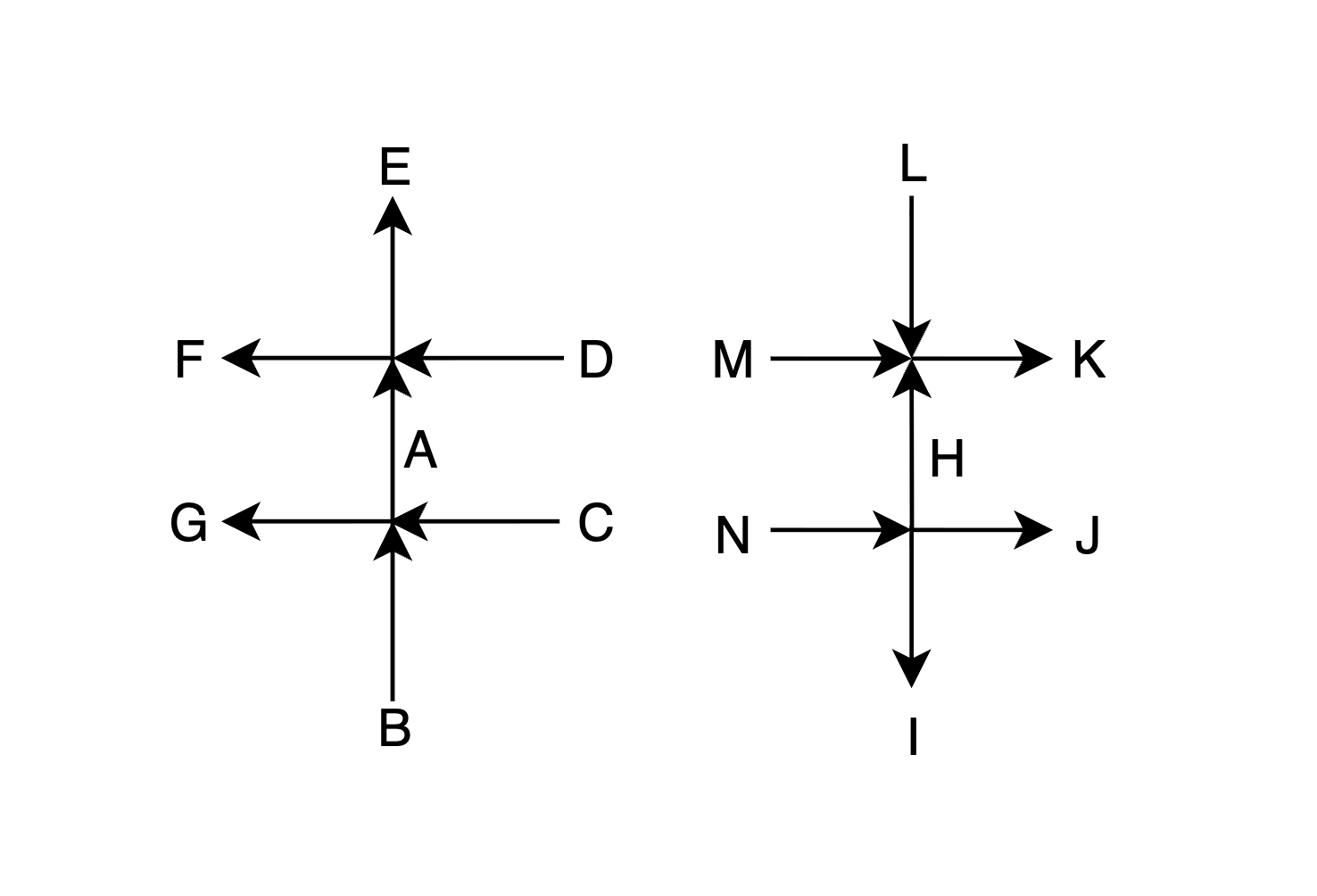}
\caption{\label{streetRep} All qualitatively different street segment relations}
\end{center}
\end{figure}

In our application to street networks we relax the condition for the dipoles
to be straight lines.
The streets segments have unique names 
(e.g. $A$, ... ,  $N$ in figure \ref{streetRep}).
In a pairwise manner
all streets segments meeting at a crossing have a spatial relation 
which can by qualitatively expressed
by a ${\cal DRA}_{f}$ base relation. 
Two thirds of the relations are already contained in ${\cal DRA}_{c}$.
The relations between the street segments are the following:
$A \;{\rm bsef}\; B$,
$A \;{\rm rser}\; C$,
$A \;{\rm rele}\; D$,
$A \;{\rm efbs}\; E$,
$A \;{\rm ells}\; F$,
$A \;{\rm slsr}\; G$,
$H \;{\rm sbsb}\; I$,
$H \;{\rm srsl}\; J$,
$H \;{\rm errs}\; K$,
$H \;{\rm fefe}\; L$,
$H \;{\rm lere}\; M$,
$H \;{\rm lsel}\; N$

\clearpage

\section{Experiments}

In our Graph-RAG application we use geographic data from OpenStreetMap.
Typically a street with a name in OpenStreetMap consists of several
street segments which are mapped on individual dipoles.
So the Hafenweg from our introductory example consists of three segments.
We refer to these segments as Hafenweg1, Hafenweg2, and Hafenweg3.
In general a uniquely named street is a sequence of $n$ $(n \geq 1)$ 
street segments $s_1, \ldots, s_i, \ldots, s_n$  .
The spatial relation between adjacent segments $s_i$ and $s_{i + 1}$ is
$s_i   \;{\rm efbs}\; s_{i + 1}$.

A verbalization of an individual named street starts with segment $s_1$.
The first linguistic pattern is $name (s_1)$ "begins at the intersection with" $name(r_i)$.
In this configuration $r_i$ would be the street segment that is part of the street crossing at which street $s$ starts.

In the appendix we provide the verbalization of a dipole relation knowledge graph that corresponds to a small area of Hamburg, Germany.

In total, 240 individual trials were conducted across various LLM configurations to find out whether the inclusion of qualitative geographic context helps improve the navigation performance of LLMs.
We are first going to provide an overview of the results obtained from our experiments, before breaking down the results by different factors such as test city or Large Language Model tested.


\begin{table}[h!]
\centering
\begin{tabular}{l c c c c}
\hline
\textbf{Group} & \textbf{\# Experiments} & \textbf{\# Successful} & \textbf{\# Failed} & \textbf{Success Rate (\%)} \\
\hline
Control Group & 120 & 0  & 120 & 0\% \\
Test Group    & 120 & 75 & 45  & 62.5\% \\
\hline
\end{tabular}
\caption{Overview of experiment results in control and test groups with success and failure counts and complimentary success rates.}
\end{table}


All trials could be executed successfully by the methods described above.
This means that we got a valid LLM response for each navigation task under all test conditions.
However, this does not imply that all responses were correct in terms of navigation success, but simply that we were able to collect all the necessary data.
In total, 240 navigation tasks were answered by the LLMs and manually labeled as either correct or incorrect.
Table 1 summarizes the overall results of our experiments.


There is a visible difference in navigation task success rates between the control and test groups.
In each group, the exact same 120 navigation tasks were executed.
Out of these 120 tasks, no trial in the control group could be labeled as successful.
Consequently, all trials in this group were labeled as failures, leading to a task success rate of 0\%.
In the test group however, 75 out of 120 trials could be labeled as successful.
With 45 remaining failures, this results in a task success rate of 62.5\%.


\begin{figure}[h!]
    \centering
    \includegraphics[width=0.8\textwidth]{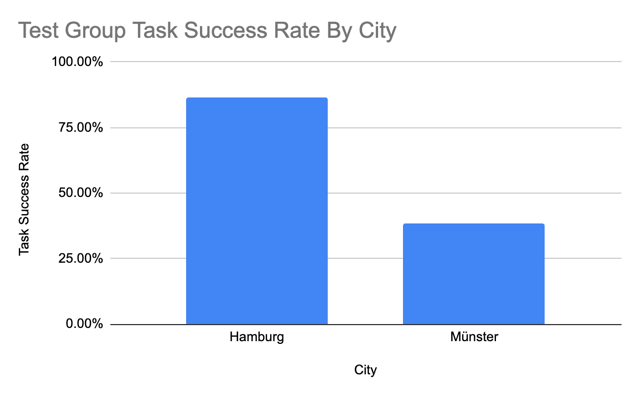} 
    \caption{Test Group Performance by City.}
    \label{fig:graphic7}
\end{figure}

When comparing the test group performance across the two test cities, we can make several key observations.
First, the task success rate was increased in both cities compared to the control group (which had a success rate of 0\% in both cities).
In Hamburg, the task success rate climbed to 86.6\% while in Münster it reached just 38.3\%. We assume that the reason or this difference is a difference in size of both data sets.
The Münster data set contains 128 individual streets and the average distance for the routes is 1272 Meters. The Hamburg data set contains 38 individual streets and the average distance for the routes is 899 Meters.
So it seems like our method gets less successful with longer routes.
We will investigate this in more detail in the future.


\begin{figure}[h!]
    \centering
    \includegraphics[width=0.8\textwidth]{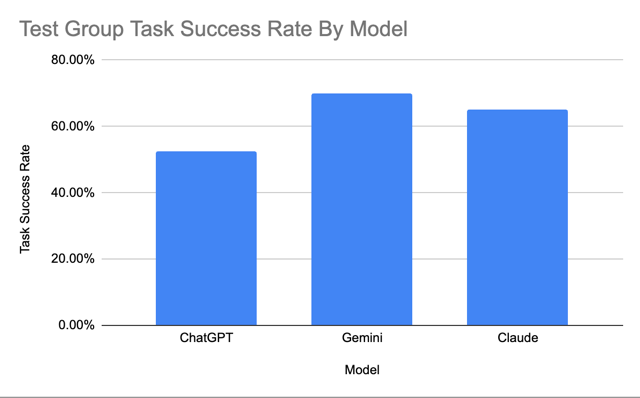} 
    \caption{Test Group Performance by LLM.}
    \label{fig:graphic8}
\end{figure}

Further, comparing the test group performance across the three tested LLMs as depicted in Figure 8, there are noteworthy differences in their performance increase.
While all models performed better compared to their control group performance of 0\%, the model which performed best across all trials in the test group was Google's Gemini 2.5 Pro.
It achieved a task success rate of 70\%, answering 28 out of 40 navigation tasks correctly.
The next best performing model was Claude 4.5, which achieved a slightly lower task success rate of 65\%.
Claude answered 26 out of 40 navigation tasks correctly.
In total, the worst performing model in the test group was GPT-4o, with 21 correct answers out of 40 possible, amounting to a task success rate of 52.5\%.

\section{Summary}

In summary, our experiments revealed that LLM navigation perfomance with additional qualitative geographic context was higher than without it.
This effect was not limited to a specific city or model, but could be observed across all combinations tested in this study.

\bibliographystyle{plain}
\bibliography{dipol}

\section*{Appendix}

\begin{verbatim}

=== Agathe-Lasch-Weg ===
Agathe-Lasch-Weg begins at the intersection with Holmbrook, Paul-Ehrlich-Straße.
Emkendorfstraße then branches off to the right.
Roosens Weg then branches off to the right.
Halbmondsweg then branches off to the left.
Klein Flottbeker Weg then branches off to the right.
Reventlowstraße then branches off to the right.

=== Ansorgestraße ===
Ansorgestraße begins at the intersection with Emkendorfstraße, Liebermannstraße.
Roosens Weg then branches off to the right.

=== Behringstraße ===
Behringstraße begins at the intersection with ['Walderseestraße', 'Behringstraße'].

=== Bernadottestraße ===
Bernadottestraße begins at the intersection with Droysenstraße.
Halbmondsweg then branches off to the right.
Halbmondsweg then branches off to the right.
Poppes Weg then branches off to the right.
Roosens Weg then branches off to the left.
Schlagbaumtwiete then branches off to the right.
Elbblöcken then branches off to the right.

=== Borchlingweg ===
Borchlingweg begins at the intersection with Ansorgestraße, Halbmondsweg.
Stindeweg then branches off to the right.
Langmaackweg then branches off to the right.

=== Corinthstraße ===
Corinthstraße begins at the intersection with Bernadottestraße.
Elbblöcken then branches off to the right.
Elbchaussee then branches off to the right.

=== Droysenstraße ===
Droysenstraße begins at the intersection with Jungmannstraße.
Preußerstraße then branches off to the left.

=== Elbblöcken ===
Elbblöcken begins at the intersection with Bernadottestraße.
Leipoldstieg then branches off to the right.

=== Elbchaussee ===
Elbchaussee begins at the intersection with Schlagbaumtwiete.
Corinthstraße then branches off to the left.

=== Emkendorfstraße ===
Emkendorfstraße begins at the intersection with Agathe-Lasch-Weg.
Röpers Weide then branches off to the left.
Lavaterweg then branches off to the left.
Reventlowstraße then branches off to the right.

=== Ernst-August-Straße ===
Ernst-August-Straße begins at the intersection with Liebermannstraße.
Roosens Weg then branches off to the left.

=== Halbmondsweg ===
Halbmondsweg begins at the intersection with Bernadottestraße.
Bernadottestraße then branches off to the left.
Ansorgestraße then branches off to the left.
Borchlingweg then branches off to the right.
Agathe-Lasch-Weg then branches off to the right.
Klein Flottbeker Weg then branches off to the left.
Reventlowstraße then branches off to the right.

=== Hammerichstraße ===
Hammerichstraße begins at the intersection with Jungmannstraße.
Walderseestraße then branches off to the left.

=== Holmbrook ===
Holmbrook begins at the intersection with Agathe-Lasch-Weg, Paul-Ehrlich-Straße.

=== Jungmannstraße ===
Jungmannstraße begins at the intersection with Reventlowstraße.
Preußerstraße then branches off to the left.
Droysenstraße then branches off to the left.

=== Klein Flottbeker Weg ===
Klein Flottbeker Weg begins at the intersection with Agathe-Lasch-Weg, Halbmondsweg, Reventlowstraße.
Zickzackweg then branches off to the left.
Droysenstraße then branches off to the left.

=== Langmaackweg ===
Langmaackweg begins at the intersection with Borchlingweg.

=== Lavaterweg ===
Lavaterweg begins at the intersection with Emkendorfstraße.

=== Leipoldstieg ===
Leipoldstieg begins at the intersection with Elbblöcken.

=== Liebermannstraße ===
Liebermannstraße begins at the intersection with Ansorgestraße, Emkendorfstraße.
Ernst-August-Straße then branches off to the right.
Bernadottestraße then branches off to the right.

=== Lobsienweg ===
Lobsienweg begins at the intersection with Droysenstraße.

=== Meistertwiete ===
Meistertwiete begins at the intersection with Bernadottestraße.

=== Olshausenstraße ===
Olshausenstraße begins at the intersection with Reventlowstraße.
Reventlowstraße then branches off to the right.
Walderseestraße then branches off to the left.

=== Paul-Ehrlich-Straße ===
Paul-Ehrlich-Straße begins at the intersection with Agathe-Lasch-Weg, Holmbrook.

=== Poppes Weg ===
Poppes Weg begins at the intersection with Bernadottestraße.
Ansorgestraße then branches off to the left.

=== Preußerstraße ===
Preußerstraße begins at the intersection with Jungmannstraße.
Droysenstraße then branches off to the right.

=== Reventlowstraße ===
Reventlowstraße begins at the intersection with Olshausenstraße.
Walderseestraße then branches off to the left.

=== Roosens Park ===
Roosens Park begins at the intersection with Ansorgestraße.

=== Roosens Weg ===
Roosens Weg begins at the intersection with Ansorgestraße.
Agathe-Lasch-Weg then branches off to the left.
Reventlowstraße then branches off to the right.

=== Röpers Weide ===
Röpers Weide begins at the intersection with Walderseestraße.
Emkendorfstraße then branches off to the right.

=== Schlagbaumtwiete ===
Schlagbaumtwiete begins at the intersection with Elbchaussee.
Bernadottestraße then branches off to the right.

=== Slevogtstieg ===
Slevogtstieg begins at the intersection with Elbblöcken.

=== Stindeweg ===
Stindeweg begins at the intersection with Borchlingweg.

=== Taxusweg ===
Taxusweg begins at the intersection with Halbmondsweg.

=== Walderseestraße ===
Walderseestraße begins at the intersection with Reventlowstraße.
Reventlowstraße then branches off to the right.

=== Zickzackweg ===
Zickzackweg begins at the intersection with Klein Flottbeker Weg.

=== Zypressenweg ===
Zypressenweg begins at the intersection with Halbmondsweg.

=== ['Walderseestraße', 'Behringstraße'] ===
['Walderseestraße', 'Behringstraße'] begins at the intersection with Behringstraße.
Walderseestraße then branches off to the right.

=== Övelgönner Hohlweg ===
Övelgönner Hohlweg begins at the intersection with Elbchaussee, Halbmondsweg.




\end{verbatim}

\end{document}